\documentclass{article}

\usepackage[preprint]{neurips_2026}

\usepackage[utf8]{inputenc}
\usepackage[T1]{fontenc}
\usepackage{hyperref}
\usepackage{url}
\usepackage{booktabs}
\usepackage{amsfonts}
\usepackage{nicefrac}
\usepackage{microtype}
\usepackage{xcolor}
\usepackage{graphicx}
\graphicspath{{figures/}}
\usepackage{wrapfig}
\usepackage{multirow}
\usepackage{makecell}
\usepackage{colortbl}
\usepackage{amssymb}
\usepackage{pifont}
\usepackage{amsmath}
\usepackage{xspace}
\usepackage{tikz}
\usetikzlibrary{positioning,arrows.meta,calc}
\usepackage[capitalize,noabbrev]{cleveref}

\definecolor{blond}{rgb}{0.98, 0.94, 0.75}

\newcommand{\cmark}{\ding{51}}%
\newcommand{\xmark}{\ding{55}}%
\newcommand{\ours}{PrimeKG-CL\xspace}

\title{\ours: A Continual Graph Learning Benchmark\\on Evolving Biomedical Knowledge Graphs}

\author{%
  Yousef A. Radwan\\
  Technology, Innovation, Entrepreneurship Department\\
  King Abdullah University of Science and Technology\\
  \texttt{yousef.radwan@kaust.edu.sa}
  \And
  Yao Li\\
  School of Computing and Information Systems\\
  The University of Melbourne\\
  \texttt{yao.li5@student.unimelb.edu.au} \\
  \And
  Qing Qing\\
  College of Computer Science and Technology\\
  Jilin University\\
  \texttt{qingqing25@mails.jlu.edu.cn} \\
  \And
  Ziqi Xu\\
   School of Computing Technologies \\
   RMIT University \\
  \texttt{ziqi.xu@rmit.edu.au} \\
  \And
  Xingtong Yu\\
  Department of Systems Engineering and Engineering Management\\
  The Chinese University of Hong Kong\\
  \texttt{xtyu@se.cuhk.edu.hk} \\
  \And
  Jiaxing Huang \\
   Department of Data Science and Artificial Intelligence \\
   Hong Kong Polytechnic University \\
   \texttt{jiaxing.huang0508@outlook.com} \\
  \And
  Renqiang Luo\thanks{Corresponding authors.} \\
  College of Computer Science and Technology\\
  Jilin University\\
  \texttt{lrenqiang@jlu.edu.cn} \\
   \And
   Xikun Zhang\footnotemark[1] \\
   School of Computing Technologies \\
   RMIT University \\
   \texttt{xikun.zhang@rmit.edu.au} \\   
}

\begin{document}

\maketitle

\begin{abstract}
Biomedical knowledge graphs underwrite drug repurposing and clinical decision support, yet the upstream ontologies they depend on (GO, HPO, MONDO, CTD) update on independent cycles that add millions of edges and deprecate hundreds of thousands more between releases. Yet existing continual graph learning (CGL) has been studied almost exclusively on synthetic random splits of static, generic KGs, a regime that cannot reproduce the asynchronous, structured evolution real biomedical KGs undergo.
To this end, we introduce \textbf{PrimeKG-CL}, a CGL benchmark built from nine authoritative biomedical databases (129K+ nodes, 8.1M+ edges, 10 node types, 30 relation types) with two genuine temporal snapshots ($t_0$ June 2021, $t_1$ July 2023; 5.83M edges added, 889K removed, 7.21M persistent), 10 entity-type-grouped tasks, multimodal node features (BiomedBERT text, Morgan fingerprints, R-GCN structural), and a per-task persistent/added/removed test stratification. On three tasks (biomedical relationship prediction, entity classification, KGQA), we evaluate six CL strategies across four KGE decoders, plus LKGE, an LLM-RAG agent, and CMKL.
We find that decoder choice and continual learning strategy interact strongly: no single strategy performs best across all decoders, and mismatched combinations can significantly degrade performance (e.g., EWC improves ComplEx by +167\% but Distillation reduces RotatE by 57\%).
Moreover, only DistMult exhibits a clear separation between persistent and deprecated knowledge ($\approx 11\times$ higher MRR on persistent vs.\ removed triples), indicating that standard metrics conflate retention of still-valid facts with failure to forget outdated ones; this effect is absent under RotatE.
In addition, multimodal features improve entity-level tasks by up to 60\%, and a recent CKGE framework (IncDE) failed to scale to our 5.67M-triple base task across five attempts up to 350\,GB RAM. Data, pipeline, baselines, and the stratified split are released openly. Dataset: https://huggingface.co/datasets/yradwan147/PrimeKGCL | Code: https://github.com/yradwan147/primekg-cl-neurips2026
\end{abstract}

\section{Introduction}
\label{sec:intro}

Biomedical knowledge graphs (KGs) underwrite drug discovery, disease understanding, and clinical decision support~\citep{Chandak2023, Huang2024}, yet they are far from static: upstream ontologies (GO, HPO, MONDO, CTD) update asynchronously, continuously introducing new entities and deprecating outdated associations. A model trained on a 2021 snapshot will face substantial new knowledge by 2023, raising the fundamental question: \emph{how can graph-based models continually learn from evolving biomedical KGs without catastrophically forgetting?}

Continual graph learning (CGL) has emerged to address this~\citep{Daruna2021}. Methods such as Elastic Weight Consolidation (EWC)~\citep{Kirkpatrick2017}, Experience Replay~\citep{Rolnick2019}, and frameworks such as LKGE~\citep{Cui2023} mitigate forgetting in knowledge graph embeddings (KGE). But existing CGL benchmarks construct temporal tasks via \emph{synthetic} random splits of static generic KGs (LKGE on FB15k-237/WN18RR; IncDE~\citep{Liu2024IncDE} on the same), missing the real dynamics: asynchronous additions from independent upstream sources, removals as ontologies are refined, and domain-specific patterns of change (GO annotations dominate additions; disease-phenotype links churn as HPO is refined) that random partitioning cannot reproduce.

In this work, we introduce \ours (\textbf{Prime}KG for \textbf{C}ontinual \textbf{L}earning), the first biomedical-scale CGL benchmark with genuine temporal evolution. We reconstruct two snapshots of PrimeKG~\citep{Chandak2023} by re-querying upstream databases: $t_0$ (June 2021) and $t_1$ (July 2023, rebuilt from nine freely accessible databases). The $t_0\!\to\!t_1$ partition exposes 5.8M added, 889K removed, and 7.2M persistent edges; the pipeline supports extension to additional snapshots. The benchmark ships with multimodal node features (BiomedBERT text, Morgan fingerprints~\citep{Rogers2004}, R-GCN structural), enabling modality-specific forgetting analysis. All structural baselines (Naive, Joint, EWC, ER, SI~\citep{Zenke2017}, Distillation~\citep{Hinton2015}, MIR~\citep{Aljundi2019}, LKGE) operate on KGE backbones; CMKL integrates all three modalities through MoE fusion.

Across six CL methods $\times$ four KGE decoders on three tasks, we uncover three key findings that are not observable in prior synthetic benchmarks. \emph{First}, decoder and CL strategy interact \emph{qualitatively}: no single strategy performs best across all decoders, and mismatched combinations can substantially degrade performance. For example, EWC improves ComplEx by +167\% and slightly benefits RotatE (+5\%), but Distillation reduces RotatE by 57\% and replay-based methods drop ComplEx by 38\%. Overall, RotatE~\citep{Sun2019} achieves the strongest performance (Naive AP $=0.084$ filtered MRR, 45\% higher than DistMult-Naive's $0.058$), and EWC+RotatE is the best configuration ($0.088$). \emph{Second}, the $t_0\!\to\!t_1$ temporal partition reveals a clear separation between persistent and deprecated knowledge that standard metrics (AP/AF) obscure. Under DistMult, naive drift produces an $\approx 11\times$ gap between persistent and removed triples, indicating that the model retains still-valid knowledge while down-ranking deprecated facts (correctly unlearning deprecated triples). In contrast, RotatE shows little such separation, suggesting that outdated knowledge is retained alongside valid facts. Finally, multimodal features improve entity-level tasks by up to 60\%, while a recent CKGE framework (IncDE) fails to scale to our 5.67M-triple base task, despite five attempts with up to 350\,GB RAM.

\noindent\textbf{Contributions.} (1) \ours, the first CGL benchmark on a real biomedical KG with genuine $t_0\!\to\!t_1$ evolution (129K+ nodes, 8.1M+ edges, 10 tasks), with multimodal features and a persistent/removed/added test stratification. (2) A 6$\times$4 CL$\times$decoder matrix + CMKL + LLM-RAG, across three tasks, 5 seeds, 5 CL metrics. (3) The decoder--CL contraindication and decoder-conditional correct-forgetting findings. (4) Open release of data, pipeline, baselines, and stratified split.

\begin{table*}[!t]
\centering
\caption{Comparison of \ours with existing continual graph learning benchmarks. \ours is the first to offer real temporal evolution on a biomedical KG with multimodal features and multiple evaluation tasks. ``Mod.'' = modality types provided (S: structural, T: textual, V: visual, M: molecular).}
\label{tab:comparison}
\small
\resizebox{0.98\textwidth}{!}{%
\begin{tabular}{lccccccccc}
\toprule
\textbf{Benchmark} & \textbf{Domain} & \textbf{Real Temp.} & \textbf{Mod.} & \textbf{\#Entities} & \textbf{\#Edges} & \textbf{\#Rel.} & \textbf{\#Tasks} & \textbf{\#Methods} & \textbf{Eval Tasks} \\
\midrule
LKGE~\citep{Cui2023} & General & \xmark & S & 15K & 310K & 237 & 5 & 6 & LP \\
DiCGRL~\citep{Daruna2021} & Robotics & \cmark & S & 1K & 10K & 9 & 4 & 4 & LP \\
IncDE~\citep{Liu2024IncDE} & General & \xmark & S & 15K & 310K & 237 & 5 & 5 & LP \\
ICEWS~\citep{GarciaDuran2018} & Political & \cmark & S & 13K & 386K & 256 & -- & -- & LP \\
MRCKG~\citep{Li2026MRCKG} & General & \xmark & S,T,V & 15K & 99K & 279 & 5 & 18 & LP \\
\midrule
\rowcolor{blond}
\textbf{\ours} & \textbf{Biomedical} & \cmark & \textbf{S,T,M} & \textbf{129K+} & \textbf{8.1M+} & \textbf{30} & \textbf{10} & \textbf{10} & \textbf{LP, KGQA, NC} \\
\bottomrule
\end{tabular}%
}
\end{table*}

\section{Related Work}
\label{sec:related}

\paragraph{Continual Learning for Knowledge Graphs.}
Continual learning sequentially learns from a stream of tasks without catastrophically forgetting prior knowledge~\citep{Kirkpatrick2017}.
LKGE~\citep{Cui2023} introduced a lifelong KG embedding framework applying EWC, experience replay, and distillation to TransE, DistMult, and RotatE~\citep{Bordes2013, Yang2015, Sun2019} on synthetic splits of FB15k-237 and WN18RR; DiCGRL~\citep{Daruna2021} evaluates on a robot manipulation KG; IncDE~\citep{Liu2024IncDE} uses incremental distillation; Synaptic Intelligence~\citep{Zenke2017} tracks importance online. More recent methods include FastKGE~\citep{Liu2024FastKGE} (LoRA adapters), ERPP~\citep{Yang2025ERPP} (evolutionary relation path passing), SAGE~\citep{SAGE2025} (scale-aware embedding), DebiasedKGE~\citep{Zhu2025DebiasedKGE} (disentangled learning), and workshop work on EWC for KG link prediction~\citep{Jhajj2025EWC_KG_CL}. A recent study~\citep{SIGIR2025CKGE} rethinks CKGE benchmarks and finds that synthetic partitioning obscures pattern shifts between snapshots, directly motivating our entity-type-grouped task design. All existing continual KGE methods are evaluated on synthetically partitioned generic KGs, leaving their effectiveness on real-world biomedical KG evolution untested. Broader continual graph learning~\citep{Zhang2022CGLB} targets citation or social networks with node or graph classification, not KG-specific tasks. Concurrent work~\citep{Li2026MRCKG} constructs multimodal continual KG benchmarks from DB15K, MKG-W, and MKG-Y using synthetic temporal splits; these general-domain datasets have $\leq$15K entities and no real temporal dynamics.

\paragraph{Biomedical Knowledge Graphs.}
PrimeKG~\citep{Chandak2023} integrates 20+ databases into a unified schema with 129K+ nodes across 10 biomedical entity types. TxGNN~\citep{Huang2024} leverages PrimeKG for therapeutic use prediction. Hetionet~\citep{Himmelstein2017} is an earlier integrative biomedical network. BioMedKG~\citep{Dang2025BioMedKG} enriches PrimeKG with biological sequences and textual descriptions, and PrimeKGQA~\citep{Yan2024PrimeKGQA} provides question-answer pairs for biomedical QA. None has been studied in a continual learning setting with real temporal evolution. Temporal KGs such as ICEWS~\citep{GarciaDuran2018} model event-based knowledge with explicit timestamps; in contrast, biomedical KGs undergo \emph{structural} evolution (entities and relation types are added or removed as scientific understanding advances), which \ours captures.

\paragraph{Multimodal Knowledge Graph Embedding.}
Multimodal KGE combines structural, textual, and visual features: MMKGR~\citep{Zheng2023MMKGR} (gated cross-modal attention), MCLEA~\citep{Lin2022MCLEA} (contrastive multimodal alignment), MoSE~\citep{Zhao2022} (score-level ensemble), and OGM-GE~\citep{Peng2022} (gradient modulation). Wu et al.~\citep{Wu2022} identify the ``greedy modality'' phenomenon where dominant modalities suppress others. These works operate in the static setting; \ours enables studying multimodal KGE under continual learning, where modality-specific forgetting patterns emerge.

\paragraph{Knowledge Graphs and Language Models.}
Recent work highlights the complementarity of KGs and large language models (LLMs)~\citep{Pan2024}. Our benchmark includes a RAG baseline~\citep{Lewis2020} combining KG retrieval with Qwen2.5-7B for continual KGQA.

In contrast to prior CGL benchmarks, \ours provides the first evaluation framework on a real-world biomedical knowledge graph with genuine temporal evolution and multimodal node features, encompassing three evaluation tracks and ten methods across two KGE decoder families.

\section{Benchmark Construction}
\label{sec:benchmark}

\begin{figure}[t]
\centering
\includegraphics[width=0.95\linewidth]{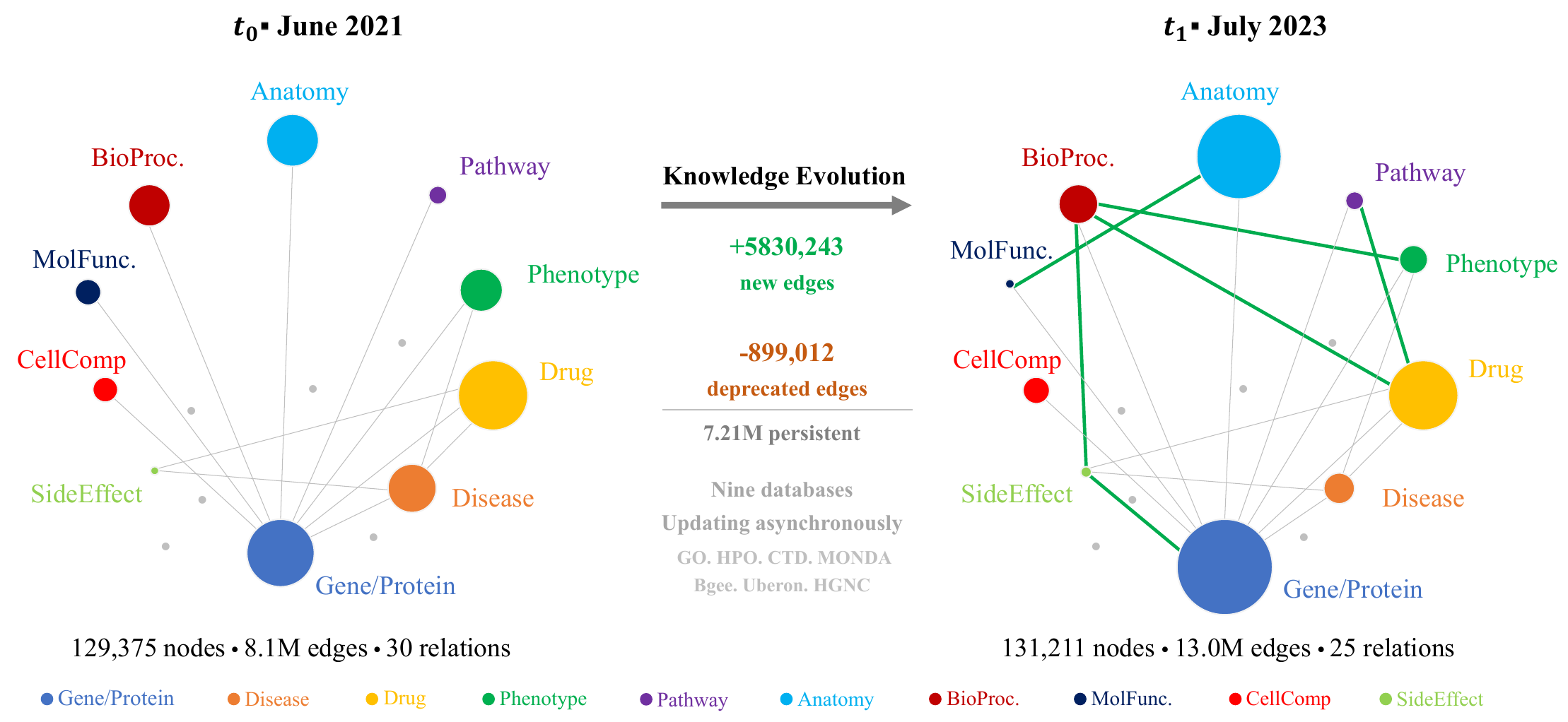}
\caption{\ours knowledge evolution from $t_0$ (June 2021) to $t_1$ (July 2023): 5.83M new edges, 889K deprecated, 7.21M persistent. Node size encodes per-entity-type frequency at each snapshot.}
\label{fig:knowledge_evolution}
\end{figure}

Two years of real biomedical knowledge, 5.83\,M new edges from nine asynchronously updating databases, 889\,K deprecated edges as ontologies were corrected, and 7.21\,M persistent edges, produces a temporal structure that random splits of FB15k-237 simply do not have. \ours captures this structure and turns it into a continual-learning workload (\cref{fig:knowledge_evolution}).

\subsection{Temporal Snapshots}
\label{sec:snapshots}

We construct two temporal snapshots of PrimeKG~\citep{Chandak2023}.
\textbf{$t_0$ (June 2021)} is the original PrimeKG release (129{,}375 nodes, 10 entity types, 8{,}100{,}498 edges, 30 relation types; integrating 20+ databases including DrugBank, DisGeNET, GO, HPO, Reactome, MONDO).

\textbf{$t_1$ (July 2023)} is reconstructed by re-querying nine freely accessible databases (Bgee, CTD, GO, Gene2GO, HPO, HPOA, MONDO, Uberon, HGNC), which update asynchronously at independent cadences (GO monthly, HPO quarterly, MONDO continuously); the resulting snapshot has 134{,}211 nodes and 13{,}041{,}729 edges across 25 relation types. Seven databases with restrictive licensing (DrugBank, UMLS, DrugCentral, SIDER) or API changes (DisGeNET) were not re-queried, so relation types sourced from these carry-forward databases have zero turnover by construction; the 5 relation types that drop from $t_1$ are drug-protein and drug-drug interactions whose source vocabulary was deprecated, and the temporal-evolution analysis (\S\ref{sec:temporal_diff}) and stratified evaluation (\cref{tab:stratified}) operate on the freshly re-queried subgraph. The pipeline supports extension to additional snapshots; per-snapshot statistics are in \cref{tab:dataset_stats}.

\begin{table}[t]
\centering
\caption{Dataset statistics for the two temporal snapshots in \ours. The temporal difference reveals substantial real-world knowledge evolution between June 2021 and July 2023.}
\label{tab:dataset_stats}
\small
\begin{tabular}{lcccc}
\toprule
\textbf{Statistic} & \textbf{$t_0$ (June 2021)} & \textbf{$t_1$ (July 2023)} & \textbf{Difference} \\
\midrule
Nodes & 129,375 & 134,211 & +4,836 \\
Edges & 8,100,498 & 13,041,729 & +4,941,231 \\
Node types & 10 & 10 & 0 \\
Relation types & 30 & 25 & $-5$ \\
\midrule
Added edges & -- & -- & +5,830,243 \\
Removed edges & -- & -- & $-889,012$ \\
Persistent edges & -- & -- & 7,211,486 \\
\midrule
Continual tasks & \multicolumn{3}{c}{10 (entity-type grouping)} \\
Evaluation tracks & \multicolumn{3}{c}{LP, KGQA, Node Classification} \\
\bottomrule
\end{tabular}
\end{table}

\subsection{Temporal Difference}
\label{sec:temporal_diff}

The temporal difference between $t_0$ and $t_1$ reveals substantial real-world knowledge evolution:
\begin{itemize}
    \item \textbf{Added edges:} 5,830,243 new triples appear in $t_1$ that were absent in $t_0$, reflecting newly discovered associations (e.g., new gene-GO annotations, updated disease-phenotype links).
    \item \textbf{Removed edges:} 889,012 triples present in $t_0$ are absent from $t_1$, representing deprecated or corrected knowledge (e.g., retracted gene-disease associations, updated ontology hierarchies).
    \item \textbf{Persistent edges:} 7,211,486 triples remain unchanged across both snapshots, providing a stable knowledge backbone.
\end{itemize}

This temporal difference is fundamentally different from synthetic CGL benchmarks that randomly partition a static graph.
The additions and removals follow domain-specific patterns: Gene Ontology annotations account for the largest share of added edges (reflecting the rapid expansion of functional annotations), while disease-phenotype associations show high turnover as HPO and MONDO ontologies are refined.

\paragraph{Stratified evaluation on added, removed, and persistent edges.}
The $t_0 \to t_1$ partition directly supports stratified continual-learning analysis. In particular, the test split of the base task ($1{,}620{,}099$ triples) stratifies cleanly into $1{,}443{,}243$ \emph{persistent} triples (89.1\%, present in both snapshots, on which a correctly updating model should \emph{retain} its predictive ability) and $176{,}856$ \emph{removed} triples (10.9\%, present in $t_0$ but deprecated in $t_1$, on which an ideal model should \emph{unlearn} its initial learned association). The later tasks provide the \emph{added} stratum: $t_1$-only triples grouped by entity type. We release this three-way stratification (\texttt{test\_stratification.json}) with the benchmark so that future work can report per-stratum MRR, quantifying correct forgetting of deprecated knowledge alongside the standard AP/AF/BWT/REM metrics.

\subsection{Continual Task Sequence}
\label{sec:task_sequence}

We organize the continual learning task sequence using an \emph{entity-type grouping} strategy.
Rather than randomly partitioning triples into artificial time steps (as in LKGE~\citep{Cui2023}), we group edges by their head and tail entity types to create semantically coherent tasks.
Specifically, we define 10 continual tasks based on the dominant entity-type pairs in PrimeKG:
\begin{enumerate}
    \item Gene/Protein $\leftrightarrow$ Gene/Protein (protein-protein interactions)
    \item Gene/Protein $\leftrightarrow$ Biological Process
    \item Gene/Protein $\leftrightarrow$ Molecular Function
    \item Gene/Protein $\leftrightarrow$ Cellular Component
    \item Gene/Protein $\leftrightarrow$ Disease
    \item Gene/Protein $\leftrightarrow$ Pathway
    \item Disease $\leftrightarrow$ Phenotype
    \item Drug $\leftrightarrow$ Disease (drug indications and contraindications)
    \item Drug $\leftrightarrow$ Side Effect
    \item Anatomy $\leftrightarrow$ Gene/Protein (tissue expression)
\end{enumerate}

\begin{figure}[t]
\centering
\includegraphics[width=\linewidth]{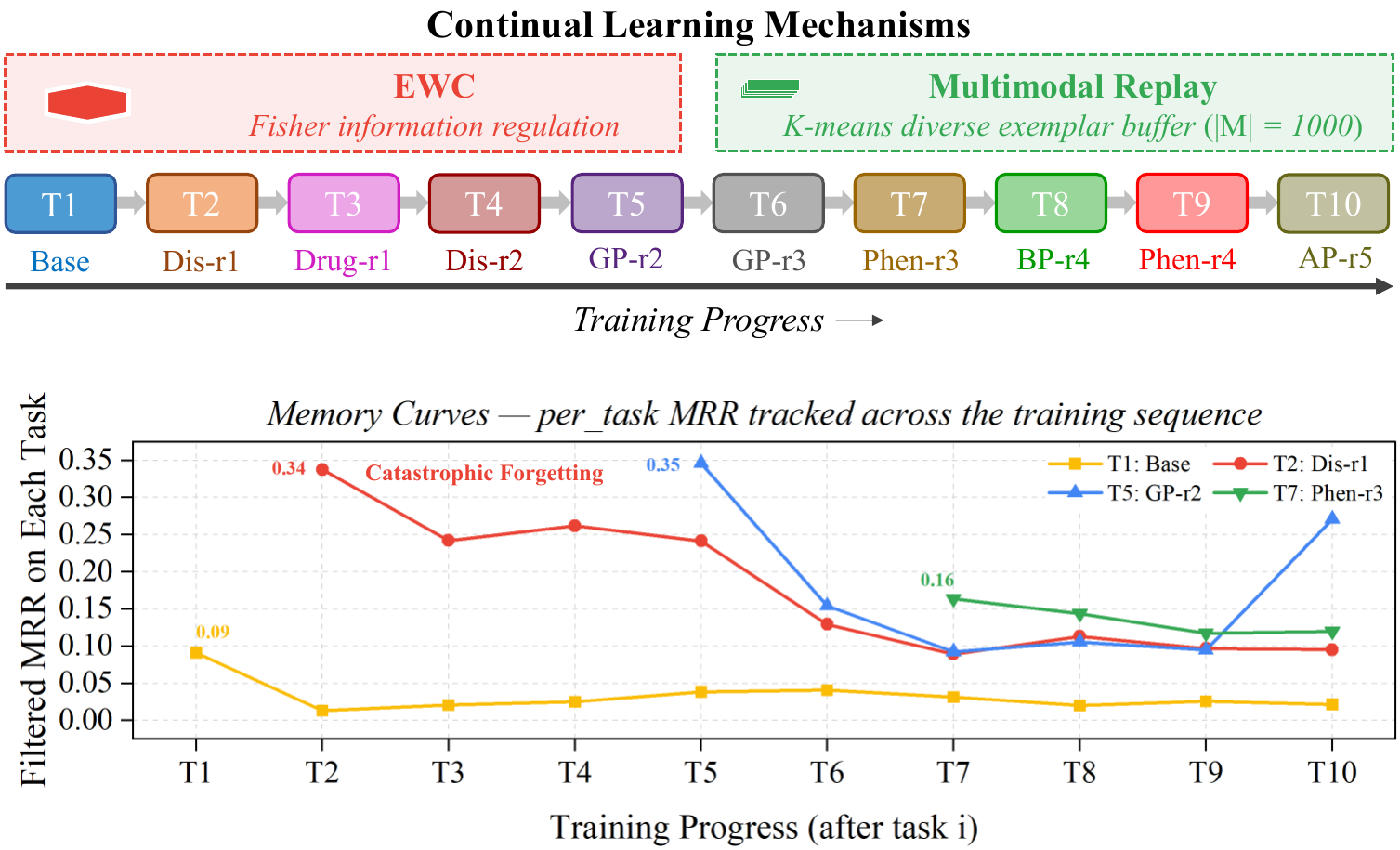}
\caption{The 10-task continual-learning sequence on \ours (top) and per-task memory curves for CMKL+DistMult under EWC + multimodal replay ($|M|{=}1000$) (bottom). Disease-r1 peaks at MRR $\approx 0.32$ when first learned (task 2) and degrades to $\approx 0.09$ by task~10, a $3.4\times$ collapse despite EWC and replay mitigations.}
%\vspace{-3mm}
\label{fig:cl_pipeline}
\end{figure}

Each task is presented sequentially at both $t_0$ and $t_1$ (\cref{fig:cl_pipeline}). The cost of getting this wrong is concrete and large: even the strongest multimodal model on this benchmark (CMKL with DistMult) sees Disease-r1 peak at MRR $\approx 0.32$ when first learned and decay to $\approx 0.09$ by task~10, a $3.4\times$ collapse on exactly the relations a clinical decision-support system would query. The grouping mirrors how upstream databases push updates (GO functional annotations, HPO phenotype links, DrugBank interactions), creating a dual challenge of \emph{task identity shifts} and \emph{temporal adaptation} ($t_0\!\to\!t_1$ within each task), analogous to domain-incremental CL in vision~\citep{vandeVen2019}.

\subsection{Evaluation Tasks}
\label{sec:eval_tasks}

\ours supports three evaluation tracks, chosen so that a continual-learning failure on each one corresponds to a concrete downstream cost: a missed drug-repurposing candidate (link prediction), a stale answer to a clinician's natural-language query (KGQA), or a newly catalogued entity assigned to the wrong type (node classification).

\paragraph{Biomedical Relationship Prediction (Link Prediction, Primary).}
Given a query $(h, r, ?)$ or $(?, r, t)$, the model ranks all candidate entities; we use filtered MRR as the per-task metric $a_{j,i}$~\citep{Bordes2013}, with Hits@$K$ ($K\in\{1,3,10\}$) in the supplement. This task supports drug repurposing, target identification, and disease gene prioritization — all of which require continually incorporating new knowledge.

Letting $a_{j,i}$ be the filtered MRR on task $i$ after training through task $j$ over $T$ tasks, we report the standard CL metrics: $\text{AP}=\tfrac{1}{T}\sum_i a_{T,i}$, $\text{AF}=\tfrac{1}{T-1}\sum_{i<T}\max_j(a_{j,i}-a_{T,i})$, $\text{BWT}=\tfrac{1}{T-1}\sum_{i<T}(a_{T,i}-a_{i,i})$, and $\text{REM}=1-\text{AF}$.

\paragraph{Knowledge Graph Question Answering (KGQA).}
We evaluate continual KGQA using a RAG pipeline~\citep{Lewis2020} that retrieves subgraph context from the current snapshot for an LLM, reporting Exact Match, token-level F1, and accuracy. This track stands in for clinical decision support, where practitioners query evolving knowledge through natural language.

\paragraph{Biomedical Entity Classification (Node Classification).}
We classify each node into its PrimeKG entity type (gene/protein, disease, drug, etc.) from KG-derived structural and optional multimodal features, reporting Macro-F1 and the CL metrics. This is the graph-learning counterpart of biomedical entity recognition: instead of locating spans in text, the model must assign the correct type to each entity in an evolving graph. Structural-only baselines reach AP$\approx0.34$--$0.37$, showing the task is non-trivial without multimodal signal.

\subsection{Multimodal Features}
\label{sec:features}

\ours ships three node-feature modalities: \textbf{textual} (768-d BiomedBERT~\citep{Gu2021} [CLS] embeddings of entity descriptions), \textbf{molecular} (1024-bit radius-2 Morgan fingerprints~\citep{Rogers2004} for drug nodes with SMILES, projected through a learned MLP), and \textbf{structural} (R-GCN~\citep{Schlichtkrull2018} embeddings capturing multi-relational neighborhood context). Together they enable study of modality-specific forgetting, a direction impossible on existing structural-only CGL benchmarks.

\section{Experiments}
\label{sec:experiments}

\subsection{Experimental Setup}
\label{sec:setup}

\paragraph{Baselines.} Ten methods spanning the major CL families: \textbf{Naive Sequential} (lower bound), \textbf{Joint Training} (oracle), regularization (\textbf{EWC}~\citep{Kirkpatrick2017}, \textbf{SI}~\citep{Zenke2017}), replay (\textbf{Experience Replay}~\citep{Rolnick2019}, \textbf{MIR}~\citep{Aljundi2019}), \textbf{Distillation}~\citep{Hinton2015}, the architecture-plus-distillation framework \textbf{LKGE}~\citep{Cui2023}, a \textbf{RAG} agent (Qwen2.5-7B)~\citep{Lewis2020}, and \textbf{CMKL} (R-GCN + DistMult, MoE fusion~\citep{Zhao2022}, modality-aware EWC, multimodal replay). Each applicable method is run with four KGE decoders: TransE~\citep{Bordes2013}, DistMult~\citep{Yang2015}, ComplEx~\citep{Trouillon2016}, RotatE~\citep{Sun2019}, via PyKEEN. Per-method details and methods we could not scale (FastKGE~\citep{Liu2024FastKGE}, ERPP~\citep{Yang2025ERPP}, SAGE~\citep{SAGE2025}, IncDE~\citep{Liu2024IncDE} which exceeded 350\,GB RAM on our 5.67M-triple base task) are in \cref{sec:supp_baselines}.

\paragraph{Setup.} $d=256$, $\eta=0.001$, batch 512, 100 epochs/task, Adam; 70/10/20 train/valid/test per task fixed across methods and seeds. EWC $\lambda=10$; replay buffer 1{,}000 triples, K-means diverse selection; CMKL uses the same buffer size, MoE fusion, per-modality Fisher. $\lambda$-sweep over $50\times$ and buffer 500--5{,}000 produced $<$10\% AP variation. Per-decoder hyperparameters are shared across CL strategies (we flag decoder-specific CL re-tuning as future work). NVIDIA V100 (32\,GB), 5 seeds $\{42,123,456,789,1024\}$, mean $\pm$ s.d., paired $t$-tests at $p<0.05$.

\subsection{Biomedical Relationship Prediction (Link Prediction) Results}%\vspace{-1mm}
\label{sec:lp_results}

\Cref{tab:main_results} reports filtered MRR over all 129K+ candidate entities. Contrary to the assumption that CL strategy dominates such a matrix, the column choice (the KGE decoder) moves AP by $20\times$, while the six CL strategies within any given column move it by at most $4\times$ and in opposite directions across columns. The remainder of this section unpacks this interaction.

\begin{table}[t]
\centering
\caption{Continual biomedical relationship prediction on \ours: AP = filtered MRR over 129K+ entities (mean $\pm$ std, 5 seeds). \textbf{Bold} = per-column best; \underline{underline} = second; \textit{italic} = oracle. AF, BWT, REM in \cref{sec:supp_results}. Discussion in \cref{sec:lp_results}.}
\label{tab:main_results}
\small
\setlength{\tabcolsep}{5pt}
\begin{tabular}{lcccc}
\toprule
\textbf{CL Strategy} & \textbf{TransE} & \textbf{DistMult} & \textbf{ComplEx} & \textbf{RotatE} \\
\midrule
Naive Sequential & $0.004 \pm 0.000$ & $0.058 \pm 0.001$ & $0.011 \pm 0.001$ & $0.084 \pm 0.005$ \\
EWC~\citep{Kirkpatrick2017} & $0.004 \pm 0.000$ & $0.058 \pm 0.001$ & $\textbf{0.029} \pm 0.002$ & $\textbf{0.088} \pm 0.003$ \\
SI~\citep{Zenke2017} & \underline{$0.005 \pm 0.000$} & $0.037 \pm 0.001$ & \underline{$0.019 \pm 0.001$} & $0.053 \pm 0.003$ \\
Distillation~\citep{Hinton2015} & \underline{$0.005 \pm 0.000$} & $0.032 \pm 0.001$ & $0.018 \pm 0.001$ & $0.036 \pm 0.000$ \\
Experience Replay~\citep{Rolnick2019} & $0.004 \pm 0.000$ & $0.051 \pm 0.001$ & $0.007 \pm 0.000$ & \underline{$0.087 \pm 0.004$} \\
MIR~\citep{Aljundi2019} & $0.003 \pm 0.000$ & $0.051 \pm 0.001$ & $0.007 \pm 0.000$ & \underline{$0.087 \pm 0.005$} \\
\midrule
LKGE~\citep{Cui2023} & $\textbf{0.039} \pm 0.001$ & --- & --- & --- \\
\midrule
Joint Training (oracle) & --- & \textit{$0.047 \pm 0.001$} & \textit{$0.025 \pm 0.001$} & \textit{$0.164 \pm 0.001$} \\
\midrule
\multicolumn{5}{l}{\emph{Multimodal (DistMult only)}} \\
CMKL (struct-only) & --- & $\textbf{0.071} \pm 0.005$ & --- & --- \\
\rowcolor{blond}
CMKL (MoE, multimodal) & --- & \underline{$0.062 \pm 0.010$} & --- & --- \\
\bottomrule
\end{tabular}

%\vspace{0.2em}
{\footnotesize RAG (Qwen2.5-7B) is decoder-free and is reported on the KGQA track only (\S\ref{sec:kgqa_results}); we omit it from this LP table.}
%\vspace{-4mm}
\end{table}

%\vspace{-1mm}
\paragraph{Per-task dynamics and the difficulty of real biomedical KGs.}
Per-task peak MRR reaches $\geq 0.25$ on several tasks (e.g., CMKL: Disease-r1 0.32, Gene/Protein-r2 0.31, Phenotype-r3 0.25), confirming meaningful patterns are learned even when AP is low. Three factors keep absolute AP modest: 129K+ candidate entities (KG-FIT~\citep{Jiang2024KGFIT} reports TransE MRR$=0.048$ on a 10K-entity PrimeKG subset), 10 entity $\times$ 30 relation types of mixed semantics, and PrimeKG sparsity ($<$1\% of possible edges) leaving many true positives outside the filter set~\citep{Sun2019}.

%\vspace{-1mm}
\paragraph{RotatE dominates the decoder spectrum.}
The single largest effect in \cref{tab:main_results} is the choice of decoder: on Naive Sequential, the four decoders span $20\times$ — TransE ($0.004$) $\ll$ ComplEx ($0.011$) $<$ DistMult ($0.058$) $<$ RotatE ($0.084$, 45\% above DistMult). We attribute this to PrimeKG-CL's compositional and approximately rotational regularities (gene-gene interactions, phenotype hierarchies, protein-protein symmetries) that rotational scoring captures but translational/bilinear cannot, unlike FB15k-237/WN18RR where DistMult and RotatE typically tie. \emph{Researchers must control for decoder choice when comparing CL strategies on biomedical KGs.}

\begin{figure}[t]
\centering
\includegraphics[width=0.78\linewidth]{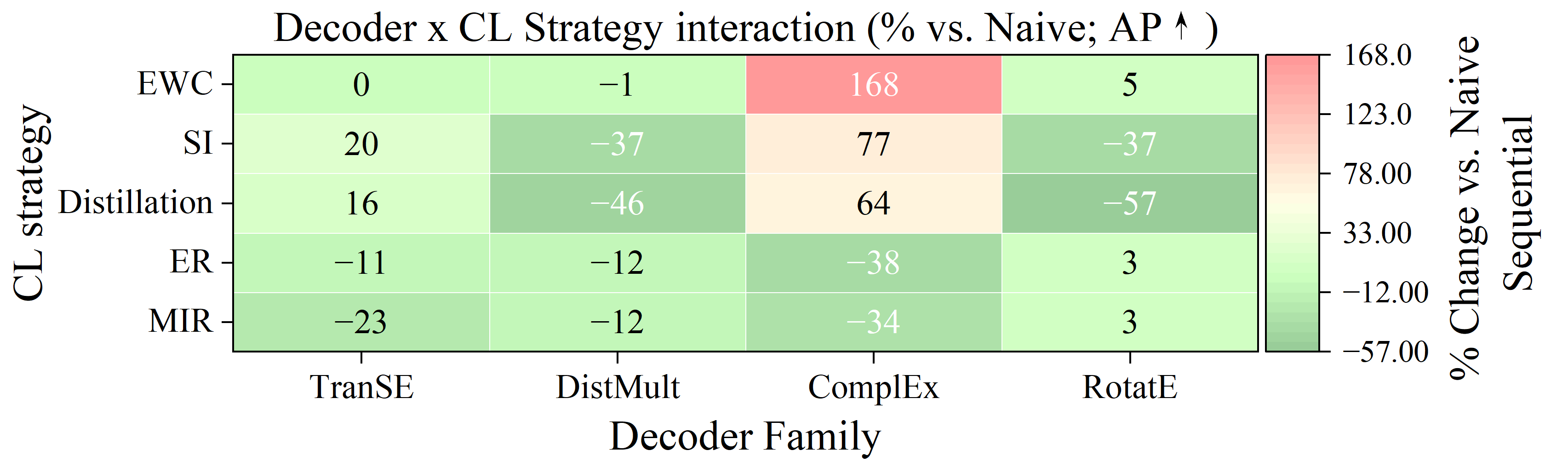}
\caption{Decoder $\times$ CL interaction (AP relative to Naive, per column): EWC transforms ComplEx (+167\%); SI/Distillation drop RotatE by 37\%/57\%; replay drops ComplEx by 38\%.}
\label{fig:decoder_cl_heatmap}
\end{figure}

%\vspace{-1mm}
\paragraph{The decoder--CL contraindication.}
The interaction between decoder family and CL strategy is the central benchmark finding: a CL strategy that helps one decoder can actively harm another, with effects large enough to erase the decoder's advantage (visualized in \cref{fig:decoder_cl_heatmap}). \textbf{EWC} is universal but decoder-sensitive in magnitude: invisible on TransE/DistMult, transformative on ComplEx ($0.011\!\to\!0.029$, +167\%, $t=34.2$, $p<10^{-4}$), and a small significant boost on RotatE ($0.084\!\to\!0.088$, +5\%, $p=0.013$), with EWC + RotatE the best configuration overall. \textbf{SI and Distillation} catastrophically degrade rotational decoders, dropping RotatE AP to $0.053$ ($-37$\%) and $0.036$ ($-57$\%) respectively while achieving AF$\approx0$ (see \cref{tab:af_matrix}): they prevent the rotational geometry from adapting. \textbf{Replay} (ER and MIR) matches EWC on RotatE ($\approx0.087$) but drops ComplEx to $0.007$, 38\% below Naive, consistent with reported ComplEx fragility under noisy negatives~\citep{Ruffinelli2020}. There is no universal best CL strategy: decoder-agnostic CL benchmarking can be actively misleading.

%\vspace{-1mm}
\paragraph{Joint Training and CMKL.}
Joint Training mirrors the contraindication: DistMult Joint reaches AP$=0.047$ ($<$ Naive's $0.058$, gradient interference across heterogeneous relations~\citep{Yu2020, Riemer2019}); RotatE Joint reaches $0.164$, nearly double the best CL configuration. The same-decoder multimodal CMKL (R-GCN+DistMult) achieves AP$=0.062\pm0.010$ ($+7$\% over DistMult-Naive/EWC at $0.058$), with AF$=0.043\pm0.008$ trading higher AP for richer drift; it is the strongest \emph{multimodal} method but not the overall best (EWC+RotatE: $0.088\pm0.003$, 42\% above CMKL-DistMult), indicating decoder and modality contribute independently.
%\vspace{-1mm}
\subsection{Biomedical Entity Classification (Node Classification) Results}
\label{sec:nc_results}
%\vspace{-1mm}
\begin{table}[t]
\centering
\caption{Continual biomedical entity classification on \ours (Macro-F1, mean $\pm$ std over 5 seeds). LKGE and RAG are excluded as not applicable to this track.}
\label{tab:nc_results}
\small
\begin{tabular}{lccc}
\toprule
\textbf{Method} & \textbf{AP} $\uparrow$ & \textbf{AF} $\downarrow$ & \textbf{BWT} $\uparrow$ \\
\midrule
Naive Sequential & $0.344 \pm 0.004$ & $0.011 \pm 0.003$ & $0.003 \pm 0.005$ \\
Joint Training & $0.370 \pm 0.002$ & $\textbf{0.003} \pm 0.004$ & $\textbf{0.022} \pm 0.004$ \\
EWC~\citep{Kirkpatrick2017} & $0.345 \pm 0.004$ & $0.008 \pm 0.003$ & $0.007 \pm 0.004$ \\
Experience Replay~\citep{Rolnick2019} & $0.344 \pm 0.006$ & $0.010 \pm 0.003$ & $0.006 \pm 0.005$ \\
SI~\citep{Zenke2017} & $0.362 \pm 0.004$ & $0.007 \pm 0.007$ & $0.015 \pm 0.006$ \\
Distillation~\citep{Hinton2015} & $0.362 \pm 0.004$ & $0.007 \pm 0.007$ & $0.015 \pm 0.006$ \\
MIR~\citep{Aljundi2019} & $0.362 \pm 0.004$ & $0.007 \pm 0.007$ & $0.015 \pm 0.006$ \\
\rowcolor{blond}
CMKL (MoE) & $\textbf{0.591} \pm 0.006$ & $0.008 \pm 0.010$ & $+0.003 \pm 0.008$ \\
\bottomrule
\end{tabular}
%\vspace{-4mm}
\end{table}

CMKL achieves the highest AP ($0.591\pm0.006$), outperforming structural-only baselines ($0.344$--$0.370$) by up to $60\%$, with near-zero forgetting (AF$=0.008\pm0.010$, BWT$=+0.003$). All methods show positive BWT, indicating that learning later associations slightly improves classification of earlier-seen entities. The structural-only baselines cluster in two groups (Naive, EWC, ER at $\approx 0.344$; SI, Distillation, MIR at $\approx 0.362$) and remain far below CMKL, confirming that the entity-classification gain is driven by multimodal features, not by CL strategy.
%\vspace{-1mm}
\subsection{Biomedical Knowledge Graph Question Answering (KGQA) Results}
\label{sec:kgqa_results}
%\vspace{-1mm}
We position the KGQA track as a stress test rather than a finished evaluation. All three configurations of Qwen2.5-7B-Instruct sit at or below token F1\,$=\,0.015$, with full RAG ($\leq 0.015$) only marginally above retrieval-only ($0.003 \pm 0.001$) and zero-shot ($0.000$). Forgetting is essentially absent (AF $\leq 0.001$) because there is little performance to forget. KGQA on a real, 129K-entity biomedical graph is a regime where current RAG pipelines are barely above retrieval; we therefore present this track as unsolved rather than saturated, calibrated to register future LLM-RAG improvements. Per-task breakdown and prompt templates are in \cref{sec:supp_kgqa}.
%\vspace{-1mm}
\subsection{Analysis}
\label{sec:analysis}
\begin{figure}[t]
\centering
\begin{minipage}[t]{\linewidth}
\centering
\includegraphics[width=\linewidth]{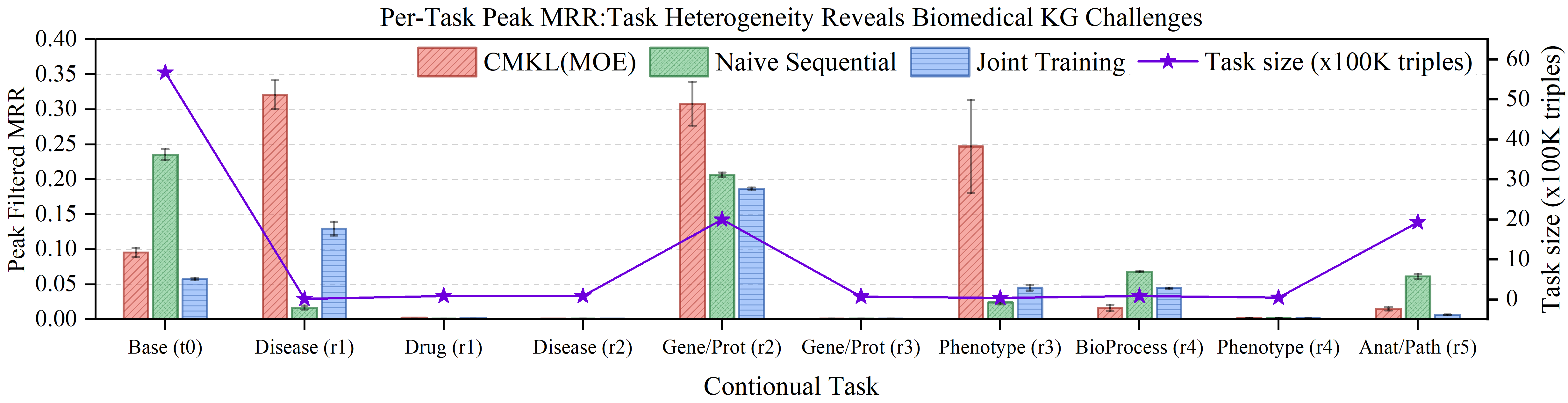}
\caption{Per-task peak MRR for CMKL, Naive, and Joint; grey line $=$ training-set size. CMKL peaks at $0.25$--$0.32$ on Disease-r1, Gene/Protein-r2, Phenotype-r3.}
\label{fig:per_task_bar}
\end{minipage}\hfill
\begin{minipage}[t]{\linewidth}
\centering
\includegraphics[width=\linewidth]{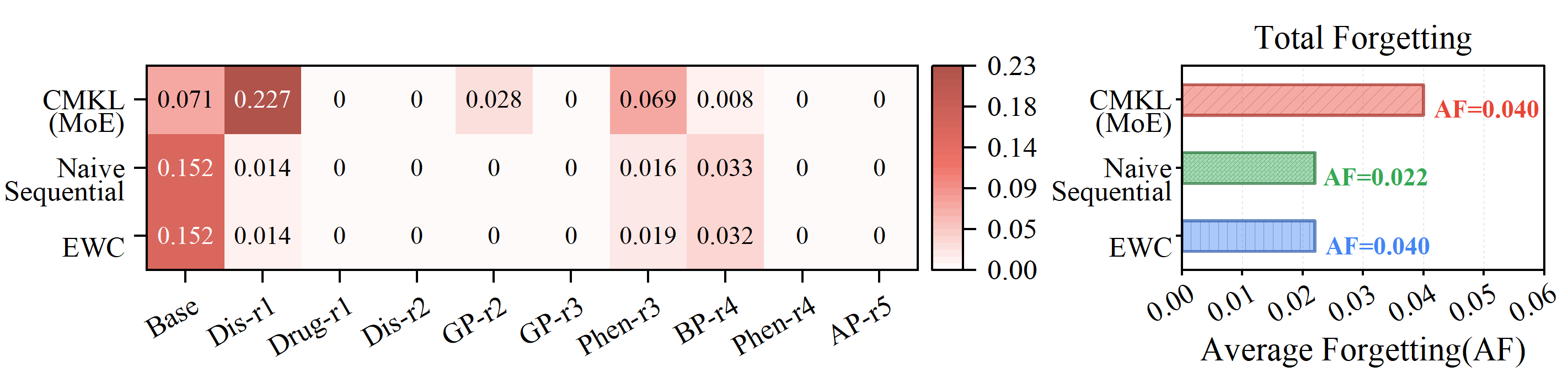}%\vspace{-2mm}
\caption{Per-task forgetting (peak $-$ final MRR). CMKL concentrates forgetting on Disease-r1 (56\%, $0.227$); Naive/EWC concentrate it on Base-$t_0$ (70\%).}
%%\vspace{-4mm}
\label{fig:modality_forgetting_heatmap}
\end{minipage}
\end{figure}

%\vspace{-1mm}
\paragraph{Per-task heterogeneity and entity-type-specific forgetting.}
Per-task peak MRR is concentrated on three tasks (\cref{fig:per_task_bar}): Disease-r1, Gene/Protein-r2, Phenotype-r3 each exceed 0.1; the others remain near zero (heterogeneous \emph{semantic} difficulty rather than scale). Forgetting is similarly concentrated (\cref{fig:modality_forgetting_heatmap}): CMKL puts 56\% on Disease-r1 (0.227), while Naive and EWC put 70\% on Base-$t_0$ (0.152) with much less Disease forgetting (0.014). Gene/protein entities make up 73\% of triples, biasing performance toward gene/protein-centric tasks.%\vspace{-1mm}

\paragraph{Correct forgetting is decoder-conditional.}
Standard AP and AF conflate two desiderata: \emph{retaining persistent} knowledge (still true at $t_1$) and \emph{unlearning deprecated} knowledge (only true at $t_0$). We split the base-task test set into $1{,}443{,}243$ persistent and $176{,}856$ removed triples, with a separate \emph{added} stratum of $1{,}156{,}876$ task-1..9 test triples newly introduced at $t_1$, and report per-stratum filtered MRR over 5 seeds in \cref{tab:stratified}. The pattern is again decoder-conditional. Under DistMult, Naive's drift yields a clean correct-forgetting signal: a persistent/removed ratio of $11\times$ ($0.096$ vs $0.009$); EWC mildly over-protects deprecated edges ($8\times$), as its uniform Fisher penalty cannot distinguish parameters that store still-true associations from those that store deprecated ones; Joint Training, which optimizes on the union of both strata, keeps them together ($2.2\times$). Under RotatE, all three methods collapse to $\approx 1.4\times$: rotational geometry retains patterns uniformly, dissolving the implicit correct-forgetting signal that DistMult-Naive achieves through drift. We release the split (\texttt{test\_stratification.json}) so future methods can target a metric that current baselines optimize only by accident.

\begin{table}[t]
\centering
\caption{Stratified filtered MRR on the final-task test set. \textbf{Persistent} triples are in both $t_0$ and $t_1$ (a correctly updating model should retain them); \textbf{removed} triples are in $t_0$ but deprecated in $t_1$ (an ideal model should forget them: lower is better); \textbf{added} triples are new in $t_1$ (mean MRR across tasks 1--9). Values are mean $\pm$ std over 5 seeds.}
\label{tab:stratified}
\small
\setlength{\tabcolsep}{4pt}
\begin{tabular}{llccc}
\toprule
\textbf{Method} & \textbf{Decoder} & \textbf{Persistent $\uparrow$} & \textbf{Removed $\downarrow$} & \textbf{Added $\uparrow$} \\
\midrule
Naive Sequential & DistMult & $0.096 \pm 0.005$ & $0.009 \pm 0.000$ & $0.055 \pm 0.001$ \\ % n=5 seeds
Naive Sequential & RotatE & $0.106 \pm 0.003$ & $0.078 \pm 0.002$ & $0.082 \pm 0.005$ \\ % n=5 seeds
\midrule
EWC & DistMult & $0.095 \pm 0.007$ & $0.012 \pm 0.001$ & $0.055 \pm 0.001$ \\ % n=5 seeds
EWC & RotatE & $0.106 \pm 0.002$ & $0.078 \pm 0.003$ & $0.086 \pm 0.004$ \\ % n=5 seeds
\midrule
Joint Training & DistMult & $0.213 \pm 0.006$ & $0.097 \pm 0.004$ & $0.081 \pm 0.002$ \\ % n=5 seeds
Joint Training & RotatE & $0.231 \pm 0.004$ & $0.160 \pm 0.003$ & $0.224 \pm 0.002$ \\ % n=5 seeds
\bottomrule
\end{tabular}
%\vspace{-3mm}
\end{table}

%\vspace{-2mm}
\section{Conclusion}
\label{sec:conclusion}
%\vspace{-2mm}
We introduced \ours, the first CGL benchmark grounded in a real biomedical KG with genuine temporal evolution (129K+ nodes, 8.1M+ edges, 10 entity-type tasks $\times$ two snapshots, multimodal features, three tasks). Across 6 CL methods $\times$ 4 decoders, decoder and CL strategy must be co-designed: EWC+RotatE is the best practical configuration (0.088), SI/Distillation drop RotatE by 37--57\%, replay drops ComplEx by 38\%, IncDE does not scale, and multimodal features add up to $+60\%$ on entity classification. Stratified evaluation reveals a decoder-conditional correct-forgetting signal that AP/AF average away. The consequences are concrete: any CL study on biomedical KGs that ignores decoder family is reporting a confound, and any study reporting only AP/AF averages over correct and incorrect forgetting. Data, pipeline, baselines, and stratification are released; \cref{sec:supp_lim,sec:supp_data} cover limitations and availability.

{\small
\bibliographystyle{unsrt}
\bibliography{refs}

@article{Li2026MRCKG,
  title={When Modalities Remember: Continual Learning for Multimodal Knowledge Graphs},
  author={Li, Linyu and Jin, Zhi and Zhang, Yichi and Jin, Dongming and He, Yuanpeng and Duan, Haoran and Luosang, Gadeng and Tashi, Nyima},
  journal={arXiv preprint arXiv:2604.02778},
  year={2026}
}

@inproceedings{Zhu2025DebiasedKGE,
  title={{DebiasedKGE}: Towards Mitigating Spurious Forgetting in Continual Knowledge Graph Embedding},
  author={Zhu, Junlin and Fu, Bo and Duan, Guiduo},
  booktitle={Proceedings of the 34th ACM International Conference on Information and Knowledge Management (CIKM)},
  year={2025},
  doi={10.1145/3746252.3761387}
}

@article{Jhajj2025EWC_KG_CL,
  title={Elastic Weight Consolidation for Knowledge Graph Continual Learning: An Empirical Evaluation},
  author={Jhajj, Gaganpreet and Lin, Fuhua},
  journal={arXiv preprint arXiv:2512.01890},
  year={2025}
}

@article{Dang2025BioMedKG,
  title={Multimodal Contrastive Representation Learning in Augmented Biomedical Knowledge Graphs},
  author={Dang, Tien and Nguyen, Viet Thanh Duy and Le, Minh Tuan and Hy, Truong-Son},
  journal={arXiv preprint arXiv:2501.01644},
  year={2025}
}

@inproceedings{Yan2024PrimeKGQA,
  title={Bridging the Gap: Generating a Comprehensive Biomedical Knowledge Graph Question Answering Dataset},
  author={Yan, Xi and Westphal, Patrick and Seliger, Jan and Usbeck, Ricardo},
  booktitle={Proceedings of the 27th European Conference on Artificial Intelligence (ECAI)},
  series={Frontiers in Artificial Intelligence and Applications},
  volume={392},
  pages={1198--1205},
  year={2024}
}

@article{Chandak2023,
  title={Building a knowledge graph to enable precision medicine},
  author={Chandak, Payal and Huang, Kexin and Zitnik, Marinka},
  journal={Scientific Data},
  volume={10},
  number={1},
  pages={67},
  year={2023},
  publisher={Nature Publishing Group}
}

@article{Huang2024,
  title={A foundation model for clinician-centered drug repurposing},
  author={Huang, Kexin and Chandak, Payal and Wang, Qianwen and Haber, Shreyas and Zitnik, Marinka},
  journal={Nature Medicine},
  volume={30},
  number={12},
  pages={3601--3613},
  year={2024},
  publisher={Nature Publishing Group}
}

@article{Himmelstein2017,
  title={Systematic integration of biomedical knowledge prioritizes drugs for repurposing},
  author={Himmelstein, Daniel Scott and Lizee, Antoine and Hessler, Christine and Brueggeman, Leo and Chen, Sabrina L and Hadley, Dexter and Green, Ari and Khankhanian, Pouya and Baranzini, Sergio E},
  journal={eLife},
  volume={6},
  pages={e26726},
  year={2017},
  publisher={eLife Sciences Publications}
}

@article{Kirkpatrick2017,
  title={Overcoming catastrophic forgetting in neural networks},
  author={Kirkpatrick, James and Pascanu, Razvan and Rabinowitz, Neil and Veness, Joel and Desjardins, Guillaume and Rusu, Andrei A and Milan, Kieran and Quan, John and Ramalho, Tiago and Grabska-Barwinska, Agnieszka and others},
  journal={Proceedings of the National Academy of Sciences},
  volume={114},
  number={13},
  pages={3521--3526},
  year={2017},
  publisher={National Academy of Sciences}
}

@inproceedings{Rolnick2019,
  title={Experience replay for continual learning},
  author={Rolnick, David and Ahuja, Arun and Schwarz, Jonathan and Lillicrap, Timothy P and Wayne, Gregory},
  booktitle={Advances in Neural Information Processing Systems},
  volume={32},
  year={2019}
}

@inproceedings{Zenke2017,
  title={Continual learning through synaptic intelligence},
  author={Zenke, Friedemann and Poole, Ben and Ganguli, Surya},
  booktitle={Proceedings of the 34th International Conference on Machine Learning},
  pages={3987--3995},
  year={2017},
  organization={PMLR}
}

@article{Hinton2015,
  title={Distilling the knowledge in a neural network},
  author={Hinton, Geoffrey and Vinyals, Oriol and Dean, Jeff},
  journal={arXiv preprint arXiv:1503.02531},
  year={2015}
}

@inproceedings{Aljundi2019,
  title={Online continual learning with maximally interfered retrieval},
  author={Aljundi, Rahaf and Caccia, Lucas and Belilovsky, Eugene and Caccia, Massimo and Lin, Min and Charlin, Laurent and Tuytelaars, Tinne},
  booktitle={Advances in Neural Information Processing Systems},
  volume={32},
  year={2019}
}

@article{vandeVen2019,
  title={Three scenarios for continual learning},
  author={van de Ven, Gido M and Tolias, Andreas S},
  journal={arXiv preprint arXiv:1904.07734},
  year={2019}
}

@inproceedings{Cui2023,
  title={Lifelong embedding learning and transfer for growing knowledge graphs},
  author={Cui, Yuanning and Wang, Yuxin and Sun, Zequn and Liu, Wenqiang and Jiang, Yiqiao and Han, Kexin and Hu, Wei},
  booktitle={Proceedings of the AAAI Conference on Artificial Intelligence},
  volume={37},
  pages={4218--4226},
  year={2023}
}

@inproceedings{Liu2024IncDE,
  title={Towards continual knowledge graph embedding via incremental distillation},
  author={Liu, Jiajun and Ke, Wenjun and Wang, Peng and Shang, Ziyu and Gao, Jinhua and Li, Guozheng and Ji, Ke and Liu, Yanhe},
  booktitle={Proceedings of the AAAI Conference on Artificial Intelligence},
  volume={38},
  number={8},
  pages={8759--8768},
  year={2024}
}

@inproceedings{Bordes2013,
  title={Translating embeddings for modeling multi-relational data},
  author={Bordes, Antoine and Usunier, Nicolas and Garcia-Duran, Alberto and Weston, Jason and Yakhnenko, Oksana},
  booktitle={Advances in Neural Information Processing Systems},
  volume={26},
  year={2013}
}

@inproceedings{Yang2015,
  title={Embedding entities and relations for learning and inference in knowledge bases},
  author={Yang, Bishan and Yih, Wen-tau and He, Xiaodong and Gao, Jianfeng and Deng, Li},
  booktitle={Proceedings of the 3rd International Conference on Learning Representations},
  year={2015}
}

@inproceedings{Sun2019,
  title={{RotatE}: Knowledge graph embedding by relational rotation in complex space},
  author={Sun, Zhiqing and Deng, Zhi-Hong and Nie, Jian-Yun and Tang, Jian},
  booktitle={Proceedings of the 7th International Conference on Learning Representations},
  year={2019}
}

@inproceedings{Trouillon2016,
  title={Complex embeddings for simple link prediction},
  author={Trouillon, Th{\'e}o and Welbl, Johannes and Riedel, Sebastian and Gaussier, {\'E}ric and Bouchard, Guillaume},
  booktitle={Proceedings of the 33rd International Conference on Machine Learning},
  pages={2071--2080},
  year={2016}
}

@inproceedings{Ruffinelli2020,
  title={You {CAN} teach an old dog new tricks! On training knowledge graph embeddings},
  author={Ruffinelli, Daniel and Broscheit, Samuel and Gemulla, Rainer},
  booktitle={Proceedings of the 8th International Conference on Learning Representations},
  year={2020}
}

@inproceedings{Schlichtkrull2018,
  title={Modeling relational data with graph convolutional networks},
  author={Schlichtkrull, Michael and Kipf, Thomas N and Bloem, Peter and van den Berg, Rianne and Titov, Ivan and Welling, Max},
  booktitle={The Semantic Web: 15th International Conference (ESWC 2018)},
  pages={593--607},
  year={2018},
  publisher={Springer}
}

@article{Gu2021,
  title={Domain-specific language model pretraining for biomedical natural language processing},
  author={Gu, Yu and Tinn, Robert and Cheng, Hao and Lucas, Michael and Usuyama, Naoto and Liu, Xiaodong and Naumann, Tristan and Gao, Jianfeng and Poon, Hoifung},
  journal={ACM Transactions on Computing for Healthcare},
  volume={3},
  number={1},
  pages={1--23},
  year={2021},
  publisher={ACM}
}

@inproceedings{Lewis2020,
  title={Retrieval-augmented generation for knowledge-intensive {NLP} tasks},
  author={Lewis, Patrick and Perez, Ethan and Piktus, Aleksandra and Petroni, Fabio and Karpukhin, Vladimir and Goyal, Naman and K{\"u}ttler, Heinrich and Lewis, Mike and Yih, Wen-tau and Rockt{\"a}schel, Tim and others},
  booktitle={Advances in Neural Information Processing Systems},
  volume={33},
  pages={9459--9474},
  year={2020}
}

@inproceedings{GarciaDuran2018,
  title={Learning sequence encoders for temporal knowledge graph completion},
  author={Garc{\'\i}a-Dur{\'a}n, Alberto and Duman{\v{c}}i{\'c}, Sebastijan and Niepert, Mathias},
  booktitle={Proceedings of the 2018 Conference on Empirical Methods in Natural Language Processing},
  pages={4816--4821},
  year={2018}
}

@article{Daruna2021,
  title={Continual learning of knowledge graph embeddings},
  author={Daruna, Angel and Gupta, Mehul and Sridharan, Mohan and Chernova, Sonia},
  journal={IEEE Robotics and Automation Letters},
  volume={6},
  number={2},
  pages={1128--1135},
  year={2021},
  publisher={IEEE}
}

@article{Pan2024,
  title={Unifying large language models and knowledge graphs: A roadmap},
  author={Pan, Shirui and Luo, Linhao and Wang, Yufei and Chen, Chen and Wang, Jiapu and Wu, Xindong},
  journal={IEEE Transactions on Knowledge and Data Engineering},
  volume={36},
  number={7},
  pages={3580--3599},
  year={2024},
  publisher={IEEE}
}

@article{Rogers2004,
  title={Extended-connectivity fingerprints},
  author={Rogers, David and Hahn, Mathew},
  journal={Journal of Chemical Information and Modeling},
  volume={50},
  number={5},
  pages={742--754},
  year={2010},
  publisher={ACS Publications}
}

@inproceedings{Wu2022,
  title={Characterizing and overcoming the greedy nature of learning in multi-modal deep neural networks},
  author={Wu, Nan and Jastrzebski, Stanis{\l}aw and Cho, Kyunghyun and Geras, Krzysztof J},
  booktitle={Proceedings of the 39th International Conference on Machine Learning},
  pages={24043--24055},
  year={2022},
  organization={PMLR}
}

@inproceedings{Peng2022,
  title={Balanced multimodal learning via on-the-fly gradient modulation},
  author={Peng, Xiaokang and Wei, Yake and Deng, Andong and Wang, Dong and Hu, Di},
  booktitle={Proceedings of the IEEE/CVF Conference on Computer Vision and Pattern Recognition},
  pages={8238--8247},
  year={2022}
}

@inproceedings{Zhao2022,
  title={MoSE: Modality split and ensemble for multimodal knowledge graph completion},
  author={Zhao, Yu and Cai, Xiangrui and Wu, Yike and Zhang, Haiwei and Zhang, Ying and Zhao, Guoqing and Jiang, Ning},
  booktitle={Proceedings of the 2022 Conference on Empirical Methods in Natural Language Processing},
  pages={10527--10536},
  year={2022}
}

@inproceedings{Lin2022MCLEA,
  title={Multi-modal contrastive representation learning for entity alignment},
  author={Lin, Zhenxi and Zhang, Ziheng and Wang, Meng and Shi, Yinghui and Wu, Xian and Zheng, Yefeng},
  booktitle={Proceedings of the 29th International Conference on Computational Linguistics},
  pages={2572--2584},
  year={2022}
}

@inproceedings{Zheng2023MMKGR,
  title={{MMKGR}: Multi-hop multi-modal knowledge graph reasoning},
  author={Zheng, Shangfei and Wang, Weiqing and Qu, Jianfeng and Yin, Hongzhi and Chen, Wei and Zhao, Lei},
  booktitle={Proceedings of the 39th IEEE International Conference on Data Engineering (ICDE)},
  pages={96--109},
  year={2023}
}

@inproceedings{Zhang2022CGLB,
  title={CGLB: Benchmark tasks for continual graph learning},
  author={Zhang, Xikun and Song, Dongjin and Tao, Dacheng},
  booktitle={Advances in Neural Information Processing Systems},
  volume={35},
  pages={13006--13021},
  year={2022}
}

@inproceedings{Liu2024FastKGE,
  title={Fast and continual knowledge graph embedding via incremental {LoRA}},
  author={Liu, Jiajun and Ke, Wenjun and Wang, Peng and Wang, Jiahao and Gao, Jinhua and Shang, Ziyu and Li, Guozheng and Xu, Zijie and Ji, Ke and Li, Yining},
  booktitle={Proceedings of the 33rd International Joint Conference on Artificial Intelligence},
  pages={2159--2167},
  year={2024}
}

@inproceedings{SAGE2025,
  title={{SAGE}: Scale-aware gradual evolution for continual knowledge graph embedding},
  author={Li, Yifei and Zhang, Lingling and Yan, Hang and Zhao, Tianzhe and Ma, Zihan and Huang, Muye and Liu, Jun},
  booktitle={Proceedings of the 31st ACM SIGKDD Conference on Knowledge Discovery and Data Mining},
  year={2025}
}

@inproceedings{SIGIR2025CKGE,
  title={Rethinking continual knowledge graph embedding: Benchmarks and analysis},
  author={Zhao, Tianzhe and Chen, Jiaoyan and Ru, Yanchi and Lin, Qika and Geng, Yuxia and Zhu, Haiping and Pan, Yudai and Liu, Jun},
  booktitle={Proceedings of the 48th International ACM SIGIR Conference on Research and Development in Information Retrieval},
  year={2025}
}

@inproceedings{Yang2025ERPP,
  title={From knowledge forgetting to accumulation: Evolutionary relation path passing for lifelong knowledge graph embedding},
  author={Yang, Jing and Jiang, Xinfa and Jiang, Xiaowen and Gao, Yuan and Yang, Laurence T and Zou, Shaojun and Yang, Shundong},
  booktitle={Proceedings of the 48th International ACM SIGIR Conference on Research and Development in Information Retrieval},
  pages={1197--1206},
  year={2025}
}

@inproceedings{Yu2020,
  title={Gradient surgery for multi-task learning},
  author={Yu, Tianhe and Kumar, Saurabh and Gupta, Abhishek and Levine, Sergey and Hausman, Karol and Finn, Chelsea},
  booktitle={Advances in Neural Information Processing Systems},
  volume={33},
  pages={5824--5836},
  year={2020}
}

@inproceedings{Riemer2019,
  title={Learning to learn without forgetting by maximizing transfer and minimizing interference},
  author={Riemer, Matthew and Cases, Ignacio and Ajemian, Robert and Liu, Miao and Rish, Irina and Tu, Yuhai and Tesauro, Gerald},
  booktitle={International Conference on Learning Representations},
  year={2019}
}

@inproceedings{Jiang2024KGFIT,
  title={{KG-FIT}: Knowledge Graph Fine-Tuning Upon Open-World Knowledge},
  author={Jiang, Pengcheng and Cao, Lang and Xiao, Cao and Bhatia, Parminder and Sun, Jimeng and Han, Jiawei},
  booktitle={Advances in Neural Information Processing Systems},
  volume={37},
  year={2024}
}
}

\newpage
\setcounter{page}{1}
\renewcommand{\thefigure}{S\arabic{figure}}
\setcounter{figure}{0}
\renewcommand{\thesection}{S\arabic{section}}
\setcounter{section}{0}
\renewcommand{\thetable}{S\arabic{table}}
\setcounter{table}{0}

\begin{center}
\Large\textbf{PrimeKG-CL: A Continual Graph Learning Benchmark on Evolving Biomedical Knowledge Graphs}\par
%\vspace{1em}
\Large Supplementary Material
\end{center}

\section{Experimental Details}
\label{sec:supp_exp}
All experiments use NVIDIA V100 GPUs (32\,GB) with PyTorch 2.0. Key hyperparameters: embedding dimension 256, learning rate 0.001, batch size 512, 100 epochs per task, 50{,}000 sampled triples per epoch. For structural-only baselines, a single EWC penalty $\lambda = 10$ is used. For CMKL, modality-specific lambdas are used: $\lambda_s = 10$ (structural), $\lambda_t = 5$ (textual), $\lambda_m = 1$ (molecular), $\lambda_f = 5$ (fusion), $\lambda_r = 50$ (main relation embeddings, set high because rotational decoders treat relations as rotation angles where any drift warps the full embedding geometry). Replay buffer: 1{,}000 triples total via K-means selection on structural embeddings. ComplEx and RotatE decoders follow the PyKEEN convention (entity dim $= 2D$, $D$ complex pairs interleaved); RotatE relations are $D$ phases initialized uniformly in $[-\pi, \pi]$, ComplEx relations are full-complex $2D$ real values initialized with Xavier. All results averaged over 5 seeds $\{42, 123, 456, 789, 1024\}$.

\paragraph{Scoring-function corrections (2026-04-12).} During a mid-revision audit we identified and fixed four bugs in the PyKEEN-based scoring pipeline that had affected earlier drafts. (i) \textbf{RotatE complex-tensor handling}: raw PyKEEN entity embeddings are stored as $[N, 2D]$ real, which must be converted to $[N, D]$ complex via \texttt{torch.view\_as\_complex}; our prior implementation treated the raw tensor as already-complex, silently corrupting the head embedding. (ii) \textbf{RotatE evaluation norm}: eval used $L_1$ instead of $L_2$ on the complex difference, inconsistent with training. (iii) \textbf{ComplEx}: the same raw-to-complex conversion bug as RotatE. (iv) \textbf{TransE training norm}: training used $L_2$ where the PyKEEN convention for TransE is $L_1$; re-running confirmed the numerical effect is within seed noise for TransE, so reported TransE numbers are unchanged. All four corrections were verified against PyKEEN's \texttt{model.score\_hrt} reference. The new RotatE and ComplEx numbers reported here post-date these fixes; we re-ran the full $4$-decoder matrix under the corrected pipeline.

\paragraph{IncDE scalability note.} We attempted to evaluate IncDE~\citep{Liu2024IncDE}, a representative incremental-distillation CKGE method, following the official repository. Over five attempts (including the bug fixes listed in the paper's GitHub issue tracker) we observed out-of-memory failures at 64\,GB, 200\,GB, and 350\,GB RAM, and deadlocks in the training loop on the 5.67\,M-triple base task. In our hands, the published implementation does not scale to the task-0 graph of \ours. We cannot rule out implementation differences relative to the authors' environment; we document our configurations alongside the released code. This negative result is informative: it highlights a gap between general-domain CKGE benchmarks ($\le$15\,K entities) and real biomedical graphs where methods that appear scalable on paper can become intractable in practice.

\section{Additional Results}
\label{sec:supp_results}
Forward Transfer (FWT) equals zero for all continual methods in our sequential setting, as tasks are non-overlapping and models are not evaluated on unseen tasks before training. Hits@1, Hits@3, and Hits@10 are available in the released evaluation code.

\begin{table}[h]
\centering
\caption{\textbf{Average Forgetting (AF) $\downarrow$} for the full 6-method $\times$ 4-decoder matrix (mean $\pm$ std, 5 seeds). Negative values mean the task improved from its first-seen-performance to end-of-sequence performance. SI and Distillation achieve near-zero AF but at a large AP cost on rotational decoders (see \cref{tab:main_results}); replay variants (ER, MIR) show moderate AF but devastating AP drops for ComplEx.}
\label{tab:af_matrix}
\small
\setlength{\tabcolsep}{4pt}
\begin{tabular}{lcccc}
\toprule
\textbf{CL Strategy} & \textbf{TransE} & \textbf{DistMult} & \textbf{ComplEx} & \textbf{RotatE} \\
\midrule
Naive Sequential & $0.020 \pm 0.000$ & $0.004 \pm 0.001$ & $0.030 \pm 0.001$ & $0.027 \pm 0.002$ \\
EWC & $0.017 \pm 0.000$ & $0.006 \pm 0.001$ & $0.010 \pm 0.001$ & $0.022 \pm 0.003$ \\
SI & $\approx 0.000$ & $-0.002 \pm 0.000$ & $-0.002 \pm 0.000$ & $0.004 \pm 0.002$ \\
Distillation & $\approx 0.000$ & $-0.002 \pm 0.000$ & $\approx 0.000$ & $-0.002 \pm 0.000$ \\
Experience Replay & $0.021 \pm 0.001$ & $0.013 \pm 0.002$ & $0.035 \pm 0.001$ & $0.027 \pm 0.002$ \\
MIR & $0.022 \pm 0.000$ & $0.013 \pm 0.001$ & $0.034 \pm 0.001$ & $0.028 \pm 0.002$ \\
Joint (oracle) & --- & $0.000$ & $0.000$ & $0.000$ \\
\bottomrule
\end{tabular}
\end{table}

\begin{table}[h]
\centering
\caption{\textbf{Backward Transfer (BWT) $\uparrow$} for the full matrix.}
\label{tab:bwt_matrix}
\small
\setlength{\tabcolsep}{4pt}
\begin{tabular}{lcccc}
\toprule
\textbf{CL Strategy} & \textbf{TransE} & \textbf{DistMult} & \textbf{ComplEx} & \textbf{RotatE} \\
\midrule
Naive Sequential & $-0.020 \pm 0.000$ & $-0.004 \pm 0.001$ & $-0.030 \pm 0.001$ & $-0.027 \pm 0.002$ \\
EWC & $-0.016 \pm 0.000$ & $-0.006 \pm 0.001$ & $-0.010 \pm 0.001$ & $-0.022 \pm 0.003$ \\
SI & $\approx 0.000$ & $+0.006 \pm 0.000$ & $+0.004 \pm 0.000$ & $+0.007 \pm 0.002$ \\
Distillation & $\approx 0.000$ & $+0.005 \pm 0.000$ & $+0.002 \pm 0.001$ & $+0.006 \pm 0.000$ \\
Experience Replay & $-0.021 \pm 0.001$ & $-0.012 \pm 0.002$ & $-0.035 \pm 0.001$ & $-0.027 \pm 0.002$ \\
MIR & $-0.021 \pm 0.000$ & $-0.011 \pm 0.001$ & $-0.034 \pm 0.001$ & $-0.028 \pm 0.002$ \\
Joint (oracle) & --- & $0.000$ & $0.000$ & $0.000$ \\
\bottomrule
\end{tabular}
\end{table}

\section{Baseline Details}
\label{sec:supp_baselines}

We evaluate the following ten methods, each with the indicated hyperparameter setting. All KGE-based baselines use embedding dimension $d = 256$, learning rate $\eta = 10^{-3}$, batch size 512, and 100 training epochs per task with the Adam optimizer.

\begin{itemize}
    \item \textbf{Naive Sequential.} Fine-tunes on each task sequentially without any forgetting-mitigation mechanism. Serves as a lower bound on knowledge retention.
    \item \textbf{Joint Training.} Trains on the union of all tasks simultaneously. Serves as a reference oracle since no forgetting can occur by construction.
    \item \textbf{EWC}~\citep{Kirkpatrick2017}. Regularization-based approach: penalizes parameter changes weighted by the diagonal Fisher information from previous tasks. Default $\lambda = 10$.
    \item \textbf{Experience Replay}~\citep{Rolnick2019}. Memory-based approach: replays a 1{,}000-triple exemplar buffer selected by K-means diverse sampling.
    \item \textbf{SI (Synaptic Intelligence)}~\citep{Zenke2017}. Regularization-based: tracks parameter importance online via accumulated gradients during training, without requiring a separate Fisher pass.
    \item \textbf{Distillation}~\citep{Hinton2015}. Knowledge-distillation approach: minimizes divergence between current-task predictions and snapshot model outputs.
    \item \textbf{MIR (Maximally Interfered Retrieval)}~\citep{Aljundi2019}. Replay variant: selects buffer exemplars most affected by the current gradient update.
    \item \textbf{LKGE}~\citep{Cui2023}. Combined architecture and distillation framework, the state-of-the-art CGL approach for KGs. Constrained to TransE backbone.
    \item \textbf{RAG Agent (Qwen2.5-7B)}~\citep{Lewis2020}. Non-parametric retrieval-augmented generation pipeline. Included for completeness; evaluated on the KGQA track.
    \item \textbf{CMKL (R-GCN + DistMult).} Multimodal CGL method with textual (BiomedBERT), molecular (Morgan FP), and structural (R-GCN) features fused via mixture-of-experts~\citep{Zhao2022}; uses modality-aware EWC ($\lambda_s{=}10, \lambda_t{=}5, \lambda_m{=}1, \lambda_f{=}5, \lambda_r{=}50$) and a 1{,}000-triple multimodal replay buffer.
\end{itemize}

\paragraph{Methods we could not run at scale.}
FastKGE~\citep{Liu2024FastKGE}, ERPP~\citep{Yang2025ERPP}, and SAGE~\citep{SAGE2025} are not included because their code was not publicly available at the time of submission. We additionally attempted IncDE~\citep{Liu2024IncDE}, following the official repository: across five attempts that scaled RAM up to 350\,GB, the published implementation did not complete training on our 5.67M-triple base task (out-of-memory failures and training-loop deadlocks). We document our exact configurations alongside the released code; the negative result itself documents the gap between general-domain CKGE benchmarks ($\le$15K entities) and real biomedical-scale graphs.

\section{Per-task Learning Matrices}
\label{sec:supp_analysis_figs}

\begin{figure}[h]
\centering
\includegraphics[width=\linewidth]{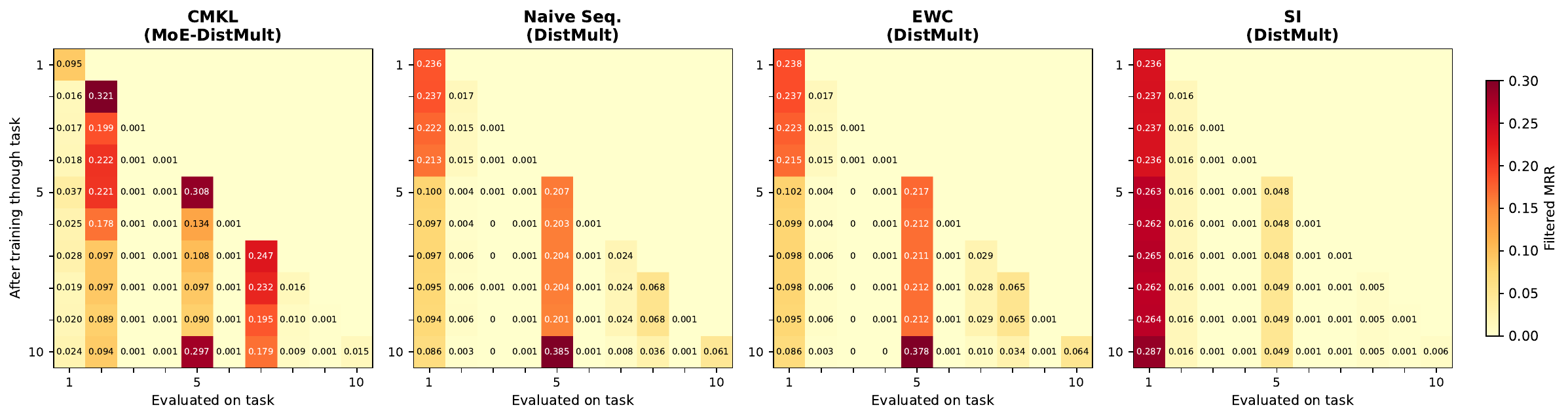}
\caption{Per-task learning matrices (DistMult, 5 seeds). Entry $(i,j)$ = filtered MRR on task $j$ after training through task $i$; diagonal $=$ peak per-task performance, off-diagonal decay $=$ forgetting trajectory. CMKL preserves the diagonal more uniformly than Naive/EWC.}
\label{fig:learning_matrix}
\end{figure}

We complement the per-task and decoder-CL visualizations in the main paper (\cref{fig:decoder_cl_heatmap,fig:per_task_bar,fig:modality_forgetting_heatmap}) with full per-task learning matrices in \cref{fig:learning_matrix}. Each cell $(i,j)$ reports filtered MRR on task $j$ after training through task $i$ (mean over 5 seeds, DistMult decoder), so the diagonal is each task's peak performance and below-diagonal columns trace catastrophic forgetting through the rest of the sequence. The matrix makes two patterns visible that the aggregate AP/AF metrics flatten: (i) Disease-r1 and Gene/Protein-r2 dominate the diagonal (peaks $\geq0.30$) but their column also shows the steepest decay, so the highest-performing tasks contribute disproportionately to forgetting; (ii) the bottom rows of the Naive matrix are nearly uniform horizontally, indicating that by the time the model has trained through later tasks, its representation no longer differentiates the easier task-types it once retained. The CMKL matrix shows the opposite pattern: peaks on Disease-r1, Gene/Protein-r2, and Phenotype-r3 are preserved through the sequence, with the dominant decay isolated to Disease-r1 alone.

\section{KGQA Details}
\label{sec:supp_kgqa}

We evaluate continual KGQA using three configurations of Qwen2.5-7B-Instruct to ablate the contribution of retrieval:
\begin{itemize}
    \item \textbf{Full RAG:} LLM with ChromaDB subgraph retrieval (200 questions/task).
    \item \textbf{Retrieval-only:} Extracts answers from retrieved triples via majority voting, without LLM generation.
    \item \textbf{Zero-shot LLM:} LLM answers directly from parametric knowledge, without any retrieval.
\end{itemize}
The zero-shot LLM achieves token F1 = 0.000 across all tasks, failing to generate correct biomedical entity names from parametric knowledge alone. Retrieval-only achieves token F1 = $0.003 \pm 0.001$, demonstrating that retrieved triples contain relevant information but simple entity extraction is insufficient. Full RAG achieves the highest performance (token F1 $\leq$ 0.015), highlighting the challenge of generating exact biomedical entity names from free-text LLM output. All configurations exhibit near-zero forgetting (AF $\leq 0.001$), as retrieval-augmented approaches store knowledge in the index rather than model weights.

\section{Analysis Details}
\label{sec:supp_analysis}

\paragraph{Per-task performance variation.}
A subset of tasks (Disease-r1, Gene/Protein-r2, Phenotype-r3) achieve substantially higher peak MRR ($>0.1$), driven by larger training sets and denser graph neighborhoods. Other tasks, including Drug (r1), Disease (r2), BioProcess (r4), and Phenotype (r4), are substantially harder, with peak MRR below 0.05 for all methods. This heterogeneity is a realistic characteristic of biomedical KGs that synthetic benchmarks with uniform random splits do not capture.

\paragraph{Gene/protein dominance effect.}
Gene/protein entities account for approximately 73\% of triples in PrimeKG, creating a substantial type imbalance. This affects all methods: performance on gene/protein-centric tasks is consistently higher than on tasks involving rarer entity types.

\paragraph{Entity-type-specific forgetting patterns.}
For CMKL (MoE), Disease-r1 accounts for 56\% of total forgetting (0.227), followed by Base ($t_0$) at 18\% and Phenotype-r3 at 17\%. In contrast, Naive Sequential and EWC concentrate forgetting on Base ($t_0$) at 70\% (0.152), with substantially less Disease forgetting (0.014). This divergence reveals that the dominant source of forgetting depends on model architecture: multimodal representations shift the vulnerability from the largest task (Base, which dominates structural embeddings) to disease-related tasks where textual and structural signals interact. Drug, Gene/Protein-r3, and Anatomy/Pathway tasks show near-zero forgetting across all methods, suggesting these entity types have more stable embedding neighborhoods.

\paragraph{Domain-specific insights.}
The biomedical domain introduces unique CGL challenges absent from generic benchmarks. First, ontology updates (e.g., MONDO disease hierarchy changes) cause systematic triple removals that look like ``forgetting'' but actually reflect corrected knowledge. Second, highly connected hub nodes (e.g., TP53, BRCA1) participate in many relation types, creating inter-task dependencies that complicate task-isolated continual learning. Third, the multimodal nature of biomedical entities means that forgetting can be modality-specific: a model may retain a drug's molecular properties while forgetting its textual description-based associations.

\section{Limitations and Future Work}
\label{sec:supp_lim}
\begin{itemize}
\item \textbf{Two snapshots.} Only $t_0$ (2021) and $t_1$ (2023) are currently included; the construction pipeline supports extension to additional snapshots as upstream databases update.
\item \textbf{Seven licensed databases excluded.} DrugBank, UMLS, DrugCentral, SIDER, and others with restrictive licensing could not be re-queried for $t_1$; including them would enrich drug-related dynamics.
\item \textbf{Gene/protein dominance.} 73\% of triples involve gene/protein entities, biasing evaluation toward this type. Type-balanced splits are a future direction.
\item \textbf{Evaluation paradigm gap.} KGE methods use filtered MRR with all-entity ranking; the RAG agent uses entity-name matching against retrieved subgraphs. Direct cross-paradigm comparison requires caution.
\end{itemize}

\paragraph{Future work.} Promising directions include integrating licensed databases for richer drug-disease dynamics; co-designed decoder-CL pairs (e.g., replay variants for Hermitian bilinear decoders, regularization schemes that preserve rotational geometry); multi-hop reasoning tasks; inductive continual learning where new entity types appear across snapshots; and foundation-model approaches for continual biomedical KG learning building on KG-LLM complementarity~\citep{Pan2024}.

\section{Data Availability, License, and Maintenance}
\label{sec:supp_data}
\textbf{License.} Code components are released under the MIT license. Data files are released under Creative Commons Attribution 4.0 International (CC BY 4.0), inheriting PrimeKG's license; users must respect upstream-database licenses where applicable. The full LICENSE file ships in the dataset archive root.
\textbf{Hosting.} Long-term hosting via HuggingFace Datasets (DOI assigned upon acceptance) and a GitHub release tag at the publication SHA. Croissant 1.0 metadata (\texttt{croissant.json}) accompanies the release.
\textbf{Maintenance plan.} We commit to (i) corrective releases for downstream-database-driven errata for at least 24 months after publication, (ii) a tagged release for each future $t_i$ snapshot we add, (iii) GitHub Issues triage with a 72h initial response target during the year following publication.
\textbf{Contents.} Temporal snapshots (\texttt{kg\_t0.csv}, \texttt{kg\_t1.csv}), task directories with 70/10/20 train/valid/test splits, multimodal features (\texttt{text\_embeddings.pt}, \texttt{mol\_features.pt}, R-GCN tensors), persistent/added/removed test stratification (\texttt{test\_stratification.json}), per-task statistics, and reference baseline + CMKL implementations.
\textbf{Datasheet for Datasets.} A datasheet following Gebru et al.\ (2021) is included as \texttt{DATASHEET.md} in the release.
\textbf{Stratified evaluation scope.} The stratified-MRR table (Table~\ref{tab:stratified}) reports 6 cells (3 methods $\times$ 2 decoders, 5 seeds each), the configurations our compute budget supported within the submission window. Extending the matrix to the full 6 CL methods $\times$ 4 decoders is one of the first follow-on experiments the released benchmark and stratification metadata enable.

%\newpage
%\input{checklist}

\end{document}